\pdfoutput=1
\documentclass[11pt]{article}

\usepackage[]{EMNLP2022}

\usepackage{times}
\usepackage{latexsym}
\usepackage[textsize=scriptsize]{todonotes}
\usepackage{url}
\usepackage{booktabs}
\usepackage{graphicx}
\usepackage{xcolor}
\usepackage{comment}
\usepackage[T1]{fontenc}
\usepackage[utf8]{inputenc}

\usepackage{microtype}

\usepackage{inconsolata}

\renewenvironment{quote}{
  \list{}{
    \leftmargin0.4cm   % this is the adjusting screw
    \rightmargin\leftmargin
  }
  \item\relax
}

\newcommand{\dataset}{{\sc DCQA}}

\title{Discourse Comprehension: A Question Answering Framework to Represent Sentence Connections}

\author{Wei-Jen Ko$^1$\ \ \ \
Cutter Dalton$^2$\ \ \ \
Mark Simmons$^3$\ \ \ \
Eliza Fisher$^4$\\ 
\textbf{Greg Durrett}$^1$\ \ \ \
\textbf{Junyi Jessy Li}$^4$\\
$^1$ Computer Science, $^4$ Linguistics, The University of Texas at Austin\\
$^2$ Linguistics, University of Colorado Boulder\\
$^3$ Linguistics, University of California San Diego\\
{\small \tt wjko@utexas.edu,
cutter.dalton@colorado.edu,
mjsimmons@ucsd.edu,
eliza.fisher@utexas.edu,}\\
{\small \tt gdurrett@cs.utexas.edu,
jessy@utexas.edu}\\
}

\begin{document}
\maketitle
\begin{abstract}

While there has been substantial progress in text comprehension through simple factoid question answering, more holistic comprehension of a discourse still presents a major challenge~\cite{dunietz2020test}.
Someone critically reflecting on a text as they read it will pose curiosity-driven, often open-ended questions, which reflect deep understanding of the content and require complex reasoning to answer~\cite{inq,tedq}.
A key challenge in building and evaluating models for this type of \emph{discourse comprehension} is the lack of annotated data, especially since collecting answers to such questions requires high cognitive load for annotators.
This paper presents a novel paradigm that enables scalable data collection targeting the comprehension of news documents, viewing these questions through the lens of discourse. The resulting corpus, \dataset{} (\textbf{D}iscourse \textbf{C}omprehension by \textbf{Q}uestion \textbf{A}nswering), captures both discourse and semantic links between sentences in the form of free-form, open-ended questions. 
On an evaluation set that we annotated on questions from~\citet{inq}, we show that \dataset{} provides valuable supervision for answering open-ended questions. We additionally design pre-training methods utilizing existing question-answering resources, and use synthetic data to accommodate unanswerable questions.
\end{abstract}

\section{Introduction}

\begin{figure}[t]
  \centering
  \includegraphics[width= 0.48\textwidth]{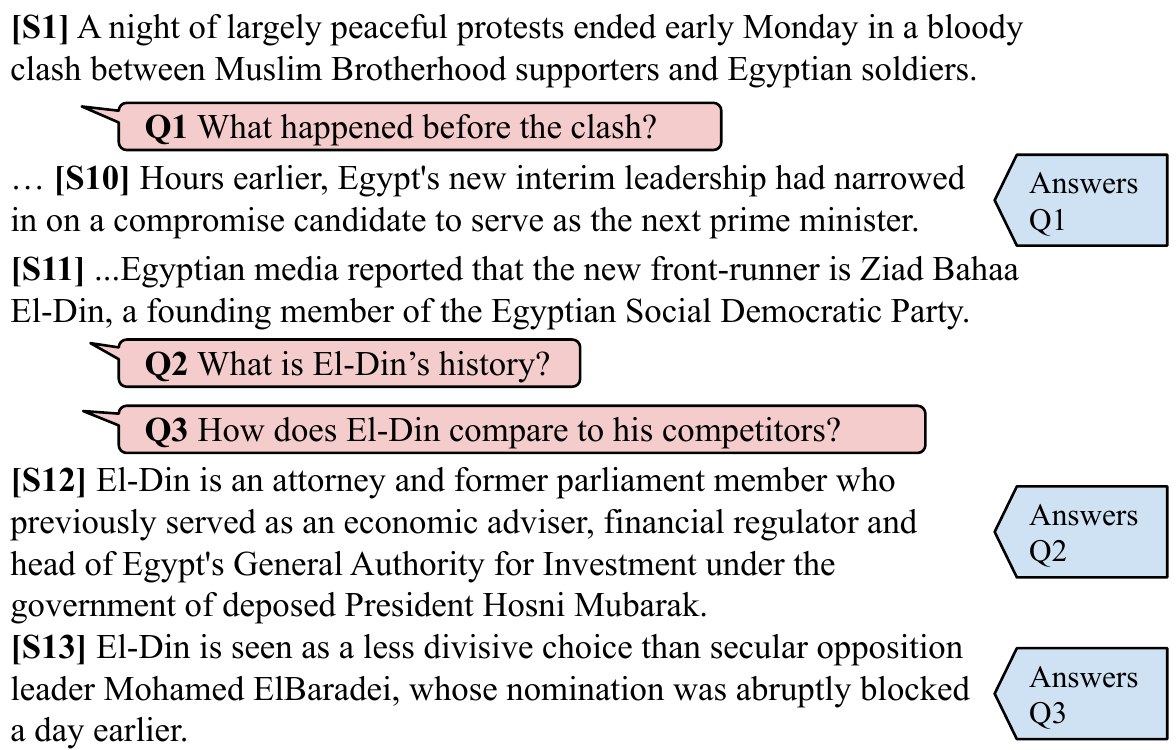}
  \caption{Discourse comprehension involves making high level inferences across sentences, often in the form of open-ended questions. This is an example from our dataset 
  \dataset. }
  \label{fig:intro-example}
\end{figure}

\begin{figure*}[t]
\vspace{-0.1in}
  \centering
  \includegraphics[scale=0.6]{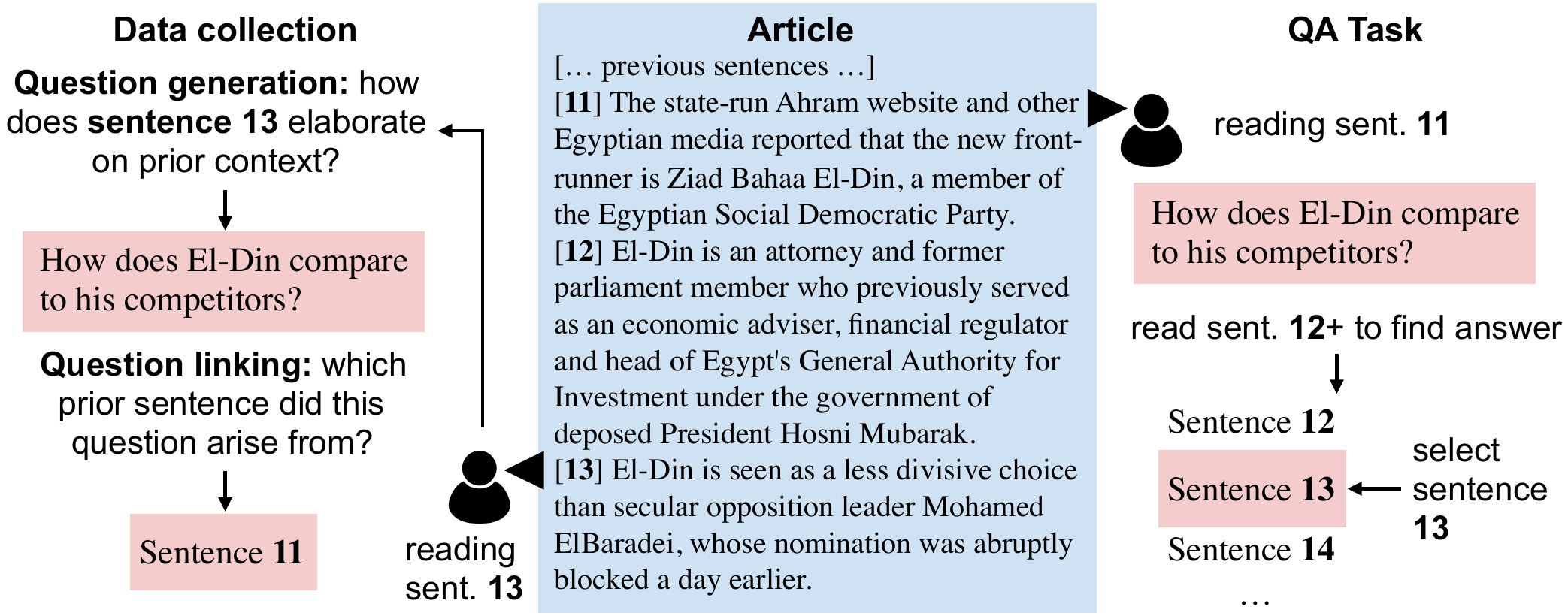}
  \caption{DCQA framework. During data collection, an annotator writes down what question a particular sentence answers and which previous sentence it arose from (i.e., its \emph{anchor point}). During QA, an annotator or an automatic system sees that question attached to the prior sentence,
  % (where someone may have spontaneously thought of it as in {\sc inquisitive}), 
  and has to find the answer in the remainder of the document.}
  \label{fig:data-collection}
\end{figure*}

Research in question answering has pushed machine comprehension to new heights, especially in answering factoid questions (e.g., SQuAD~\cite{squad} and Natural Questions~\cite{NQ}) and even questions that require multiple steps of reasoning~\cite{hotpotQA}.
Yet while existing systems are effective at sifting through a long document to find a fact, they fail to achieve what we consider \emph{discourse comprehension}.
Deep comprehension of a discourse requires establishing temporal and semantic relationships across abstract concepts in various parts of a document~\cite{hobbs1985coherence},  going beyond understanding the individual pieces of knowledge conveyed or the details of events~\cite{wegner1984comprehension}.
The gap between existing QA frameworks and discourse comprehension has been highlighted in recent work.  \citet{dunietz2020test}  emphasized the lack of narrative understanding capability in existing machine reading comprehension systems and benchmark datasets. Recent research~\cite{inq,tedq} demonstrated that human reading comprehension is marked by the spontaneous generation of \emph{open-ended}
questions \emph{anchored} in one part of an article; some of these questions are later answered in the article itself, forming a connection in the discourse (Figure~\ref{fig:intro-example}). 
\citet{inq} showed that compared to existing QA datasets, these questions are products of higher-level (semantic and discourse) processing (e.g.,``how'' and ``why'' questions) whose answers are typically complex linguistic units like whole sentences. Such reader-generated questions are far out of reach from the capabilities of systems trained on current QA datasets.

This view of discourse comprehension falls within the linguistic framework of Questions Under Discussion (QUD)~\cite{velleman2016question,de2018qud}, where discourse progresses by continuously evoking implicit or explicit questions and answering them. The open-endedness of these questions is triggered by readers' psychological mechanisms including corrections of knowledge deficits and active monitoring of the common ground~\cite{graesser1992mechanisms}. 
These fundamental differences in how questions are constructed from existing, mostly factoid QA datasets lead to difficulty spotting answers with keyword or paraphrastic overlaps. Additionally, the questions are contextualized, as they rely on an established common ground and the anchor point.

Collecting question-answer pairs that target discourse comprehension entails a high cognitive load for human annotators. Consequently, existing resources designed to train models to answer such questions are scarce. 
\citet{tedq} introduced TED-Q, a smaller dataset covering 6 TED talks, including 1102 answered questions. \citet{inq}'s {\sc inquisitive} dataset consists of more (19k) questions evoked under the QUD paradigm, yet they did not annotate answers.

We present \dataset{} (\textbf{D}iscourse \textbf{C}omprehension by \textbf{Q}uestion \textbf{A}nswering), a dataset of 22,394 English question-answer pairs distributed across 606 English news articles
(Figure~\ref{fig:data-collection}). We view \dataset{} as a key resource to train QA systems to answer discourse-driven, contextual, and open-ended questions as those in {\sc inquisitive}. On its own, \dataset{} is a discourse framework that represents connections (or relationships) between sentences via free-form questions and their answers. 

\dataset{} uses a novel crowdsourcing paradigm inspired by QUD recovery~\cite{de2018qud,riester2019constructing}: 
instead of collecting the answer labels given the question, we start from an answer sentence and collect the question. Specifically, for each sentence in an article, annotators ask a question that reflects how the main purpose of the sentence connects to an anchor sentence in prior context, such that the sentence is the answer to the question.
This paradigm is scalable while maintaining a diverse range of open-ended question types,
%~\cite{cao}; 
a varied distribution of distance between question anchor and answers, and the ability to capture interesting linguistic phenomena. We further demonstrate %(Section~\ref{sec:inquisitive_data}) 
that annotating answers given the questions is much more subjective and challenging.

We present two experiments using \dataset. We first evaluate a model's ability to extract the correct sentence to answer questions in \dataset. Our Longformer~\cite{lfm} baseline achieves 67\% accuracy, doing well in light of human performance of 72.2\%. However, the model struggles much more than humans do when the answer sentence is further away from the question anchor point.

Our main experiment evaluates whether \dataset\ can be used to train systems to answer questions such as those in {\sc inquisitive}, where the questions are asked without seeing any upcoming context that may contain answers.
To enable evaluation over {\sc inquisitive}, we collect a human-annotated test set. We find that existing datasets, including SQuAD, QuAC, ELI5, and TED-Q, yield poor accuracy (1.4\% to 25\%) on extracting answer sentences for the answerable questions in {\sc inquisitive}. In contrast, training on \dataset\ leads to a performance of 40.8\%. 
We further design ways to enable pre-training using SQuAD and ELI5, which leads to significant performance improvements. 
Finally, we present a pipeline system for question-answering that handles unanswerable questions utilizing synthetic data.
\footnote{We release \dataset{} at \href{https://github.com/wjko2/DCQA-Discourse-Comprehension-by-Question-Answering}{https://github.com/wjko2/DCQA-Discourse-Comprehension-by-Question-Answering}.}

\section{Background and related work}\label{sec:background}

\paragraph{Discourse and question answering.}

Discourse theories profile the relationships between linguistic units in a document that make it coherent and meaningful; these relations could be temporal, causal, elaborative, etc.~\cite{mann1988rhetorical,lascarides2008segmented,pdtb}. Most of these theories define fixed relation taxonomies, which have been viewed as templatic question~\cite{prasad2008discourse,pyatkin2020qadiscourse}.
In contrast, 
QUD makes use of free-form questions to represent discursive relationships \cite{riester2019constructing};
each utterance in a text builds on the reader's common ground and can evoke questions; each utterance is also an answer to an implicit (or explicit, if present) question.

{\sc inquisitive}~\cite{inq} is a collection of such curiosity-driven, QUD-style questions generated by readers as they read through a document (i.e., asking questions on-the-fly without seeing any upcoming context); specifically, 19K questions for the first 5 sentences across 1500 news documents from three sources: Newsela~\cite{newsela}, WSJ articles from the Penn Treebank~\cite{ptb}, and Associated Press~\cite{ap}.
Each question is attached to a span in the current sentence the reader is reading, e.g.,
\begin{quote}
\small
...~It's an obvious source of excitement for Doyle, who is 27 years old and has \underline{severe autism.} \\
{\bf Q:} Why is this important?
\end{quote}
\vspace{0.5em}

One major motivation of \dataset{} is to train models to answer these questions. Thus \emph{as opposed to} creating a ``challenge dataset'', we are mainly exploring more scalable ways for data collection.

TED-Q~\cite{tedq} is also a QUD dataset annotating questions and answers simultaneously over 6 TED talk transcripts. In contrast, we propose a data collection paradigm that works from the answers back to the context for lower cognitive load and higher scalability. Also, all answers from TED-Q are within 5 sentences after the question anchor, which is not the case in our dataset.

\paragraph{Open-ended question answering.}

While there are many datasets for question answering, most of %the questions in existing datasets 
them are factoid in nature~\cite{squad,NQ}. \citet{inq} found that the question types are so different that pre-training on factoid questions \emph{hurts} the performance of \emph{generating} open-ended questions. 
\citet{eli5} collected human-written answers for their open-ended ``ELI5'' questions, but the questions and answers are not grounded in a document context.
QuAC \cite{QuAC} introduces information-seeking question-answer pairs given context in dialogue; in contrast \dataset{} aims to represent discursive and semantic links that cover the full document. Also QuAC questions are conditioned on previous dialogue turns, which is not the case in \dataset{}.
\citet{nlquad} collected paragraphs with a subheading that is a question from news articles, so the whole paragraph can be seen as the answer to the question. However, more than 86\% of the time in {\sc inquisitive} (Section~\ref{sec:inquisitive_data}), and 99\% for TED-Q, the answers to naturally generated open-ended questions consist of one sentence in the text.
By more precisely locating the answers, \dataset{} also establishes finer-grained connections within a document.

\section{The \dataset\ dataset}
We present \dataset\ using a new data collection paradigm that describes how sentences in an article relate to prior context; specifically, what \emph{question} is the current sentence answering given its prior context, inspired by literature in QUD discovery~\cite{de2018qud,riester2019constructing}.

We annotate news documents due to their rich linguistic structure that mixes narratives, different lines of events, and perspectives from parties involved~\cite{van2013news,choubey2020discourse}.

\begin{figure*}[t]
  \centering
  \includegraphics[width= \textwidth]{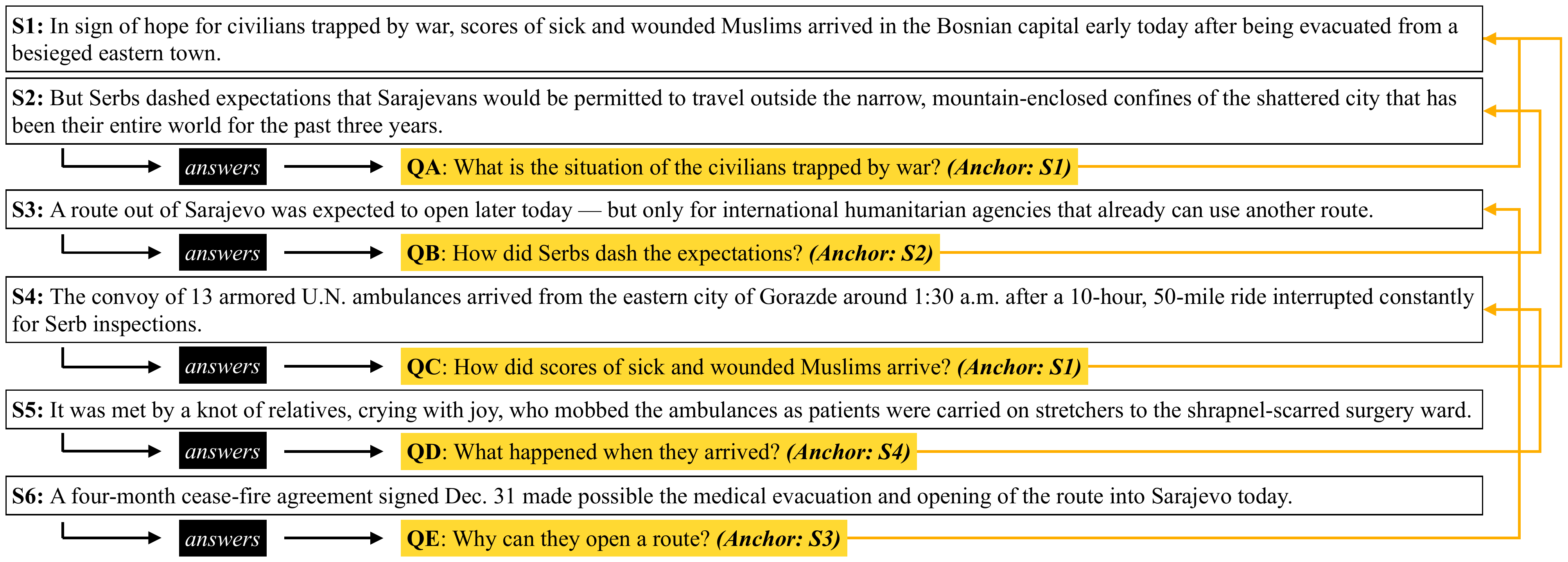}
    \caption{Snippet of an annotated article: each sentence after the first is annotated with a question linked to an earlier sentence where it arises from (which we call the question's \emph{anchor sentence}). Black arrows represent question generation; orange arrows represent question linking.}
  \label{fig:example}
\end{figure*}

\subsection{Annotation paradigm}

%\paragraph{Methodology}
We design a crowdsourcing task, illustrated in Figure~\ref{fig:data-collection}. At a high level, data collection consists of two \emph{simultaneous} tasks: \emph{question generation} given an answer sentence, and \emph{question linking} that places the question in a precise anchor point in the document. Annotators start from the beginning of the article. For each new sentence, they write a question that reflects the main purpose of the new sentence and indicates how the new sentence elaborates on prior context. The new sentence is thus the answer to the question. 
Because the question should arise from one of the previous sentences in the article, the annotator also chooses which earlier sentence most naturally evokes the question (which we refer to as the question's \emph{``anchor''} sentence).

We instructed annotators to ask open-ended questions that require some explanation as an answer, instead of questions that could be answered simply by a place/person/time or yes/no. They are also told that when writing the question, they should assume that the question could be asked and understood by people only reading the earlier sentences, following the QUD framework and ~\citet{inq,tedq}. This means avoiding reference to any information first introduced in the new sentence and avoiding copying phrases only used in the new sentence. Reusing phrases from or referencing previous sentences is allowed. 

An example of the first 6 sentences of a fully-annotated article is shown in Figure~\ref{fig:example}, where we can see each sentence after the first is annotated with a question. For instance, sentence 4 mainly describes the trip of U.N.~ambulances arriving, and it provides more detail on how the sick and wounded Muslims described in sentence 1 arrive. 

We illustrate the \textbf{Question Answering task} in the right pane of Figure~\ref{fig:data-collection}.
Given the anchor sentence and the question, the task is to extract a sentence in the upcoming context that would answer this question. Because the questions are contextual, they typically need to be associated with the anchor to be comprehensible (we illustrate the importance of identifying the anchor through model analysis in Section~\ref{sec:intrinsic}). Additionally, since we seek to answer questions that are generated during the natural progress of discourse, the questions presuppose the  common ground established by the reader, which consists of all context %from the beginning up through
up to the anchor sentence. (Because of this question generation paradigm, in \dataset{}, as well as {\sc inquisitive} and TED-Q, all answers to the questions are after the anchor sentence.) Thus the QA task is that given the context, anchor, and question, find the sentence that contains the answer. 
For example, S1 in Figure~\ref{fig:example} evokes QA and QC, and the task is to find the answer sentences 2 and 4, respectively:
\begin{quote}
\small
    \textbf{[Context+Anchor]}: In sign of hope for civilians trapped by war, scores of sick and wounded Muslims arrived in the Bosnian capital early today after being evacuated from a besieged eastern town. \\
    \textbf{[QA]}: \emph{What is the situation of the civilians trapped by war?} \\
    \textbf{[Answer to QA]}: sentence 2. \\
    \textbf{[QC]}: \emph{How did scores of sick and wounded Muslims arrive?} \\
    \textbf{[Answer to QC]}: sentence 4.
\end{quote}

\subsection{Annotators and corpus}

The data we aim to collect consists of free-form, mostly open-ended questions. The nature of the data is inherently subjective, and the free text annotation poses challenges for automatic evaluation of quality. Therefore, we take measures to both recruit good annotators and ensure annotation quality.

Our annotation team consists of three expert annotators, as well as a small number of \emph{qualified and trained} crowd workers. We first piloted the task among the expert annotators, who were undergraduate linguistics researchers familiar with linguistics literature on discourse coherence.
The crowdsourcing task was then collectively developed to be accessible to crowdworkers and to provide a list of constraints for the questions; we show in Appendix~\ref{sec:instructions} the key instructions.

We use Amazon Mechanical Turk as our crowdsourcing platform.
To ensure quality, we designed a qualification task consisting of one article. We manually inspected the quality of the responses of each candidate, and only workers who asked satisfactory questions were invited to continue with the task. Throughout the process, we also monitored the responses of the workers and reminded them about specific guidelines when necessary.
The workers are paid around \$10/hr. No demographic information was collected.

\paragraph{Corpus}
We annotate a subset of the articles used in the {\sc inquisitive} dataset~\cite{inq} (c.f.~Section~\ref{sec:background}). In contrast to {\sc inquisitive} which only contains curiosity-driven questions about the first five sentences for each article, we annotate up to the first 20 sentences. \dataset\ consists of 22,394 questions distributed across 606 articles, 51 of which are annotated by experts. 
Each article is annotated by 2 annotators.

\section{Analyzing \dataset}\label{sec:data_analysis}
This section presents a series of analytical questions we pose in order to understand \dataset\ questions, answers, and potential challenges.

\begin{table}[t!]
\centering
\small
\begin{tabular}{c|lllll}
\toprule
  Similarity&1 &2&3&4 & 5\\
  \midrule
   Percentage&26.1&14.6&17.6&16.9&24.9 \\
   Same anchor &35.8&23.7&34.8&43.1&69.8 \\
  \bottomrule
\end{tabular}
\caption{Human rating of question similarity for questions with the same answer sentence (top row), and fraction of questions with the same anchor sentence in each bucket (bottom row).}
\label{tab:similarity}
\end{table}

\paragraph{Are questions similar given the  same answer sentence?}
We first quantify the level of subjectivity in our question-answering paradigm via two angles:
%. We analyze this from two angles: 
(1) Given the same answer sentence, how similar are the questions asked by different annotators? (2) When questions are similar, do annotators agree on which sentence is the anchor?

Since the questions are free-text in form, we use human judgments to assess their similarity.
We asked our expert annotators to rate the similarity of a sample of 261 pairs of questions asked with the same answer sentence, on a scale of 1-5. Table \ref{tab:similarity} (row ``percentage'') shows that the questions different people ask may differ: while 41.8\% of the questions are highly similar with ratings of 4-5, 40.7\% of the questions are semantically different with ratings 1-2.

Table \ref{tab:similarity} (row ``same anchor'') shows how often annotators agree on the anchor sentence. When people ask the same questions (similarity 5), the percentage of agreement on the anchor sentence is high.
Similar questions can have different anchors but answered in the same sentence; Appendix~\ref{app:examples} shows an example.
Distinct questions can also share the same anchor sentence, as shown in Figure~\ref{fig:intro-example}, questions 2 and 3.

\paragraph{How well can humans do the QA task?}
%\label{sec:humananswer}
Our annotation paradigm produces open-ended questions given the answer sentence, while ultimately the question-answering task seeks to find the answer given the question. Thus we explore to what extent humans can, given a question, find the original answer sentence where the question was generated. 
We asked an expert annotator to answer questions in a randomly sampled subset of 1175 questions from the validation and test sets, without informing them of the ``gold'' answer labels. 
If the annotator thinks there are multiple answers, he could annotate multiple sentences. 
After the answers are annotated, the gold answer is shown and the annotator adjudicates the two versions of answers.  

The agreement between the annotator and the gold labels is 72.2\%. In cases where the annotator selected multiple answer sentences, we report the expected agreement when randomly choosing one of them. 
For 69.5\% of the questions, the annotator exactly chose the gold answer; for 5.3\%, the annotator selected multiple answers including the gold.
On 2.3\% of the data, the annotator thinks the gold answer cannot be used to answer the question without a stretch, thus should be regarded as noise.
For the rest, the annotator thinks that both answers are reasonable as there are multiple possible answers, although they did not originally choose the gold. In these cases, the annotator tended to choose an earlier sentence than the gold. An example for multiple answer sentences is in Appendix \ref{app:examples}.

\begin{table}[t!]
\centering
\small
\begin{tabular}{c|lll}
\toprule
  Distance&1 &2-4 & $\ge$ 5\\
  \midrule
   \dataset& 48.8& 24.2 & 27.0 \\
   TED-Q&58.2 &41.8  & \phantom{0}0.0 \\
   {\sc inquisitive}&18.1 &27.8  &54.1 \\
  \bottomrule
\end{tabular}
\caption{The distribution of the distance (in terms of sentence ID differences) between the answer and anchor sentences.}
\label{tab:distance}
\end{table}

\paragraph{How far away are answer sentences from question anchors?}
Table \ref{tab:distance} shows the distances between the question anchor and the answer sentence, by difference of sentence IDs.
While for a large proportion of questions, the answer sentence could be found just a few sentences after the question anchor sentence, there are also some questions with answers that are far away later in the article.

Compared to TED-Q, \dataset{} contains fewer questions that are answered by the immediate next sentence. 
\dataset{} also has answers $\ge$5 sentences away from the question, which TED-Q does not have.
This means that fewer questions in \dataset{} can be trivially found in the immediate next sentence.
Additionally, this also shows that the sentence connections captured by \dataset{} naturally result in a different structure from QADiscourse~\cite{pyatkin2020qadiscourse} that expresses discourse relations across adjacent sentences in templatic questions.

We also show the figures for {\sc inquisitive} over a subset that we annotate for answers (Section~\ref{sec:inquisitive_data}). Most of the {\sc inquisitive} answers are far away from where the question was asked; this is in stark contrast with TED-Q, highlighting the distributional differences between TED talks and news.

\paragraph{Linguistic characteristics of \dataset{}.}

An interesting effect of the question generation process is that annotators often find semantic links between two parts of the article that are not pragmatically or discursively connected. For example:
\vspace{0.5em}
\begin{quote}
\small
\textbf{[Context]}: This small Dallas suburb's got trouble. Trouble with a capital T and that rhymes with P and that stands for pool. More than 30 years ago, Prof. Harold Hill, the con man in Meredith Willson's 'The Music Man,' warned the citizens of River City, Iowa, against the game.\\
\textbf{[Sentence 6]}: Now kindred spirits on Addison's town council have barred the town's fanciest hotel, the Grand Kempinski, from installing three free pool tables in its new lounge.\\
\textbf{[Question]}: Just how fancy is the Grand Kempinski?\\
\textbf{[Sentence 13]}: At the lounge [of the Grand Kempinski], manager Elizabeth Dyer won't admit patrons in jeans, T-shirts or tennis shoes.
\end{quote}
\vspace{0.5em}
This question arises from the adjective \emph{fanciest} used in Sentence 6. The discursive intent of Sentence 6 is to describe the events that make the game of pool relevant to the Dallas suburb mentioned; the assertion that the Grand Kempinski hotel is the ``town’s fanciest'' is not discursively salient at this point, though it is relevant later in the article, as indicated by Sentence 13. This shows that a conscious reader was able to connect two pragmatically unrelated sentences by their semantic content. Such links have been deemed important in discourse literature, e.g., the Discourse Graphbank~\cite{wolf2005representing} and the Entity Relation between two adjacent sentences in the Penn Discourse Treebank~\cite{pdtb}. 

Finally, we present an analysis about the type of the questions asked in Appendix \ref{sec:type1}, using the schema in~\citet{cao}. \dataset{} has a good coverage of the key question types in existing high-level question answering datasets (TED-Q and {\sc Inquisitive}); the most frequent question types are concept, cause, procedural, and example.

\section{Question answering on \dataset}\label{sec:intrinsic}

As each of the questions already has a designated answer sentence during data collection, \dataset{} can be used to learn models for this type of QA, or as a testbed for existing models.

\vspace{-0.3em}
\paragraph{Model} Because our task involves reasoning over long documents,
we choose Longformer \cite{lfm} as our model. 
Longformer is a transformer model with an attention
pattern that combines a local windowed attention and global attention on pre-selected, task-dependent tokens only. This makes the computation time scale linearly with the sequence length instead of quadratic (as in conventional transformers), allowing faster computation on longer sequences, as well as better performance on QA tasks with long contexts.

Our model operates similarly to the BERT QA model for SQuAD \cite{BERT}. For the model input, we concatenate the question, the anchor sentence, and the article, separated by delimiters. Note that by passing in the full article, the model has access to all context prior to the anchor.
Since the goal of our model is to predict a sentence instead of the answer span, we add two tokens in front of each sentence in the article, the start of sentence token \texttt{[sos]} and the sentence ID. The model is trained to predict the span of the two added tokens in front of the answer sentence.

\begin{table}[t]
\small
\centering
\begin{tabular}{c|cccc}
\toprule
             & Train & Val. & Test \\ \midrule
\# questions & 20942 & 718  & 734  \\
\# docs      & 555  & 22   & 30  \\ \bottomrule
\end{tabular}
\caption{Train/test/validation splits of \dataset{}.}
\label{tab:train-test}
\end{table}

\vspace{-0.3em}
\paragraph{Settings}
We use our expert annotated question-answer pairs for the validation and test sets, and crowd annotation for the training set. Table~\ref{tab:train-test} shows the distribution of articles and questions across training, testing and validation sets.

Because we will later experiment on {\sc inquisitive} (Section~\ref{sec:extrinsic}), we exclude articles overlapping with the {\sc inquisitive} test set (10 articles containing 376 questions) during training.

We use HuggingFace~\cite{wolf-etal-2020-transformers} for our implementation. We use the Adam \cite{adam} optimizer with $(\beta_1,\beta_2)=(0.9,0.999)$ and a learning rate of 5e-5.

\vspace{-0.3em}
\paragraph{Results} The fine-tuned Longformer model achieves 67.2\% accuracy. Note that a baseline that predicts the immediate next sentence after the anchor would achieve an accuracy of 40\%. We also experimented training with TED-Q, which achieved an inferior result of 39.5\%. 

We observe the overall performance is about 5\% lower than a human's (72.2\%, Section~\ref{sec:data_analysis}). 
However, the model is doing substantially worse on answers that do not immediately follow the anchor (Table~\ref{tab:qa_intrinsic_distance}; human performance is reported on the 1,175 annotated examples in Section~\ref{sec:data_analysis}). We observe that human performance for answers further away from the anchor is also lower, reflecting that establishing longer-distance discursive connections is also more subjective for humans. 
Table \ref{tab:type} in Appendix~\ref{app:analysis:qtype} stratifies model performance per question type.

\begin{table}[t!]
\centering
\small
\begin{tabular}{c|ll}
\toprule
 Type &Model &Human  \\
  \midrule
1 &85.4&81.8\\
 2-4&46.3&69.3 \\
 $\geq$5&51.0&57.1\\
  \bottomrule
\end{tabular}
\caption{Model and human accuracy for answer sentence extraction, statified by distance between question anchor and answer.}
\label{tab:qa_intrinsic_distance}
\end{table}

\paragraph{How important are the context and anchor?}
To evaluate the influence of the context and anchor sentences, we experiment on 3 ablation settings. The results are shown in Table \ref{tab:5}.

\noindent (1) \textbf{\em No prior context:} We remove all the sentences prior to the anchor sentence from model inputs (while keeping the anchor itself). This setting resulted in an accuracy of 66.4\%, which is slightly worse yet on par with the context present, showing that the model does not use much information from the sentences before the anchor.

\noindent (2) \textbf{\em Unspecified anchor:} For the Longformer model, we do not concatenate the anchor sentence with the question, while keeping the full article intact (which includes the prior context and the anchor). We observed an accuracy of 43.5\%, which is a substantial 23.7\% drop from the version with the anchor sentence present. This shows that the model is struggling to locate the anchor if it is not given as supervision.

\noindent (3) \textbf{\em No prior context and no anchor:} In this setting, in addition to applying both modification from above: no prior context in the input article and the anchor sentence is not concatenated with the question for the model, the anchor is also removed from the input article. This yields an accuracy of 42.6\%, slightly worse but on par with (2).

We can see from (2) and (3) that specifying which sentence the question arises from is very important for the model performance, while the presupposed common ground does not play such a significant role. 
%This is because many of the questions cannot be understood when standing alone, and must be read toghether with the source sentence.

As a small experiment to assess whether humans can find the anchor points, we asked one of our expert annotators to try answering a set of 90 questions under those ablation settings. We found that for (1) and (2), the human performed almost the same as providing all the information, while (3) is 35\% worse. This shows that human are better at finding the anchor from the article while it is not specified, provided that the full article is present. Additionally, the questions are contextual, and requires the anchor to comprehend. 

%These ablation studies showed a ``challenge'' experimental setting to be pursued in future work, where the model needs to find the anchor sentence.

\begin{table}[t]
\centering
\small
\begin{tabular}{l|cc}
\toprule
            & accuracy \\ \midrule
Original model   & 67.2          \\
No prior context   & 66.4          \\
Unspecified anchor  & 43.5          \\
No context and no anchor   & 42.6          \\
\bottomrule
\end{tabular}
\caption{Results for ablating context and source sentence}
\label{tab:5}
\end{table}

\section{Question answering on external datasets} \label{sec:extrinsic}

We first describe our collection of the Inquisitive test set (Section~\ref{sec:inquisitive_data}).
Then we explore only answerable questions (Section~\ref{sec:qa_answered}), and design pre-training methods to allow the use of other QA datasets. Finally, we present a pipeline system that also handles question answerability (Section~\ref{sec:qa_full}), by predicting whether a question is answerable using synthesized data from \dataset{}.

\subsection{{\sc inquisitive} answer collection}\label{sec:inquisitive_data}

To construct the evaluation set, for a subset of {\sc inquisitive} questions in their test set, two of our expert annotators independently chose the sentence that contains the answer. Then they met to adjudicate a final version of the answers. 

This newly annotated data consists of answers to 719 {\sc inquisitive} questions from 60 articles.
Before adjudication, the annotators agree on whether there is an answer 78.8\% of the time. Out of those questions at least one annotator had answers,  42.7\% have total agreement, and another 7.3\% have partial agreement. 
This shows that answering the {\sc inquisitive} questions is challenging, even for trained linguistics senior students.

After adjudication, 424 of the questions have answers. 57 of the questions have multiple valid answers (for 30 cases, related information spreads across multiple sentences; for 27 cases, the questions can be answered from multiple angles).

During testing, we judge the model as correct if it predicts any one of the answers.

\subsection{Answering answerable questions}\label{sec:qa_answered}
We use Longformer as discussed in Section~\ref{sec:intrinsic} as our base model. Additionally, we explore the utility of other existing QA datasets via pre-training: (1) we synthetically augment SQuAD 2.0~\cite{squad} for better compatibility of data format between \dataset\ and {\sc inquisitive}; (2) we utilize ELI5~\cite{eli5} to provide more supervision for answering open-ended questions.
We also explore using TED-Q for training.
We also experiment on training with QuAC~\cite{QuAC} for comparison, since a portion QuAC questions are open-ended questions, though its domain and the dependency of the questions on previous dialog turns make the format different from ours. We do not use previous dialog turns in our experiments. 

\vspace{-0.3em}
\paragraph{SQuAD pretraining}
While
\dataset\, and {\sc inquisitive} contain  mainly high-level questions, their format is different: {\sc inquisitive} questions are asked about a sub-sentential span in the document, and the questions are more dependent on the span.

To pre-train the model to bridge this formatting gap, we create a synthetic dataset from SQuAD that simulates the input format with an annotated question span utilizing coreference substitution.  Preprocessing details are descibed in Appendix \ref{app:pretrain:squad}.

\vspace{-0.3em}
\paragraph{ELI5 pretraining}

We also sought existing data for additional supervision to answer open-ended questions.
The most related dataset we found is ELI5 \cite{eli5} for long-form question answering. While the questions in the dataset comes with a human-written answer, they do not have context to form a question answering text. Preprocessing details are described in Appendix \ref{app:pretrain:eli5}.

\vspace{-0.3em}
\paragraph{Settings}

When using either SQuAD or ELI5 pre-training, we pretrain for 6 epochs. When using both types of pretraining, we directly mix the data together, and pretrain for 3 epochs.\footnote{Also, using all SQuAD questions performed better when both types of pre-training are performed, while only using the answered questions performed the best when pre-training with SQuAD-only.} The paremeters are tuned on discourse questions validation set. All other settings are the same as in Section~\ref{sec:intrinsic}.

\begin{table}[t]
\centering
\small
\begin{tabular}{l|cc}
\toprule
            & \dataset{} & {\sc inquisitive} \\ \midrule
SQuAD   & 34.7 & 9.9         \\
ELI5    & 1.3  & 1.4         \\
QuAC & 35.9 & 17.2       \\ \midrule
TED-Q       & 39.5 & 25.0        \\
~~~~+SQuAD+ELI5 & 46.0 & 38.2       \\ \midrule

DCQA        & {\bf 67.2} & 40.8        \\
~~~~+SQuAD      & 66.7 & 41.3        \\
~~~~+ELI5       & 64.0 & 42.9        \\
~~~~+SQuAD+ELI5 & 66.0 & {\bf 45.5}        \\
~~~~~~~~+TED-Q & 64.0 & 40.6        \\
\bottomrule
\end{tabular}
\caption{Question answering results. Systems trained on \dataset{} (includes pre-training) can get the strongest performance on {\sc inquisitive}.}
\label{tab:1}
\end{table}
 
\vspace{-0.3em}
\paragraph{Results}
Table \ref{tab:1} shows for both \dataset{} and all {\sc inquisitive} questions with an answer, the accuracy values of answer sentence extraction. Training only on TED-Q, synthesized SQuAD, QuAC, and ELI5 examples
produces poor results, showing that existing datasets cannot be directly used to train model to answer the open-ended questions meant to facilitate discourse comprehension.

When tested on {\sc inquisitive}, training on \dataset{} achieves the best performance across all settings, showing its ability to provide supervision for open-ended question answering.
Although pre-training did not help for answering \dataset{} questions,
for {\sc inquisitive}, both types of pretraining are helpful individually, and using them together yields the best result. We conducted a binomial significance test between \dataset{} and S+E+\dataset, and found that the improvement is statistically significant. 
Additionally, pre-training also improved performance when used with TED-Q (rows 3 vs.\ 4). This shows that our design to adapt the two datasets is successful. TED-Q used on top of the other two pre-training did not improve performance.

\subsection{Handling question answerability}\label{sec:qa_full}
We experiment on the full open-ended question answering task: given context and question, answer the question only if there is an answer to be found in the article.

While \dataset{} does not come with unanswerable questions, we generate them by truncating parts of the articles that contain answers.\footnote{While we could in theory simply ``take out'' the answer sentence, it would leave the text incoherent and unnatural.}
Specifically, we truncate each article in the training set of \dataset{} to the first 12 sentences, and label  questions with answers after sentence 12 as unanswered.
Using this data, we train a model to predict whether a question is answerable given the text, following the setup of models in Sections~\ref{sec:intrinsic} and~\ref{sec:qa_answered}. We then combine this model with models in Section~\ref{sec:qa_answered} into a pipeline: first predict whether an answer exists, then provide the answer.
For the full task, we split the collected {\sc inquisitive} answers into a validation (223 examples) and test (496 examples) sets. We adjust the threshold of predicting on the validation set of discourse questions.

Results show that the model could predict if there is an answer correctly 84\% of the time on the synthetic discourse questions test set, but only 67\% of the time when tested on {\sc inquisitive}. In both datasets, about 59\% of the questions are answered. This shows that although our way of generating synthetic data is useful, answerability prediction is challenging.

We report the result of the pipeline system in Table \ref{tab:pipeline}, applying first our answerability model, then the corresponding question answering model. We use F1 score of correctly predicted questions following~\citet{squad}. If the model predicts no answer, we treat it as not making a prediction when calculating the F1 score. 

While \dataset{} and pre-training is useful,
the numbers clearly shows that a full open-ended question answering system, that includes answerability prediction, is an extremely challenging task. Answerability prediction   presents a bottleneck here, and we leave for future work to design better stratagies and models.

\begin{table}
\centering
\small
\setlength{\tabcolsep}{0.5em}
\begin{tabular}{l|lll}
\toprule
  System &\dataset & S+E+\dataset  & S+E+TED \\
  \midrule
  F1 & 0.260&0.358&0.242\\

\bottomrule
\end{tabular}
\caption{Pipeline system question answering results}
\label{tab:pipeline}
\end{table}

\section{Conclusion} 
We present \dataset{} that connects pieces in a document via open-ended questions and full-sentence answers. \dataset{} is collected via a new paradigm that regards the main purpose of a new sentence as an answer to a free-form question evoked earlier in the context. Consequently, this paradigm yields both discourse and semantic links across all sentences in a document.
\dataset{} is introduced with the goal of providing a more scalable data collection paradigm, also as initial resource, for answering open-ended questions for discourse comprehension. Our experiments showed that \dataset{} provides valuable supervision for such tasks.

\section*{Limitations}
\dataset{} collects questions in a reactive manner: the answer is first observed before the question is generated. This is, by design, different from methods where questions are elicited as a person reads (i.e., without seeing upcoming context, as in {\sc Inquisitive} and TED-Q). Seeing the answer before asking the question inevitably results in a slight distributional shift from datasets such as {\sc Inquisitive}, as seen in Table~\ref{tab:type2} (Appendix~\ref{sec:type1}). Qualitatively, we observe that the questions tend to be a bit more specific than {\sc Inquisitive}, and answers are more easily associated with a particular sentence.

Another notable difference is that \dataset{} does not address unanswerable questions. While we designed synthetic data augmentation methods to train models to handle such questions, this is challenging, as discussed in Section~\ref{sec:qa_full}. We hope future work could find better solutions.

Multi-sentence answers exist much more frequently in high-level question answering than factoid QA; in \dataset{}, it happens if questions elicited from different answer sentences share an anchor sentence (Appendix~\ref{app:examples}). We leave multi-sentence answers for future work.

Finally, although \dataset{} is designed as a general paradigm for data collection, the dataset presented in this paper is collected on English news articles. Thus the distribution of questions and answers may change by genre and/or language, which should be explored in future work.

\section*{Acknowledgments}
We are grateful for the feedback provided by anonymous reviewers. This work was partially supported by NSF grants IIS-2145479, IIS-2145280, and IIS-1850153. We acknowledge the Texas Advanced Computing Center at UT Austin for many of the results within this paper.

\bibliography{anthology,custom}
\bibliographystyle{acl_natbib}
%\clearpage 

\appendix

\section{Additional analysis of \dataset{}}

\subsection{Additional examples}\label{app:examples}
An example showing similar questions with different anchors but the same answer sentence:
\begin{quote}
\small
    \textbf{[Sentence 11]}  Now, those animals, once just visitors, have established resident populations -- and they are spreading. \emph{\textbf{[Q1]} How far have the Wolverines spread?}\\
    \textbf{[Sentence 12]} ``We have growing evidence of them using larger and larger areas over time,'' Aubry said. \emph{\textbf{[Q2]} How far have the Wolverines gone in their repopulation? }\\
    \textbf{[Sentence 13]} So far, scientists have confirmed resident wolverine populations from the North Cascades to as far south as this bait lure south of Highway 2 west of Leavenworth. \textbf{Answers both Q1 and Q2.}
\end{quote}

In the example below, the expert annotator found multiple answer sentences for the question whose gold answer was Sentence 12.
\begin{quote}
\small
\textbf{[Sentence 1]}: Amid skepticism that Russia's war in Chechnya can be ended across a negotiating table, peace talks were set to resume Wednesday in neighboring Ingushetia. \emph{\textbf{[Question]}: What has been the fallout of the war?}\\
\textbf{[Sentence 12 (original and expert answer)]}: The Russian offensive has turned Grozny into a wasteland littered with rotting bodies, twisted metal and debris.\\
\textbf{[Sentence 13 (expert answer)]}: Hardly a building is untouched.\\
\textbf{[Sentence 14 (expert answer)]}: The war has also cost Russia dearly -- in lives, prestige and rubles.
\end{quote}

\subsection{What questions are asked?}
\label{sec:type1}

We further examine what types of questions are asked. We fine-tune a classifier based on pretrained BERT~\cite{BERT} using the data and classification scheme from \citet{cao}.\footnote{Human evaluation (with one of our expert annotators) of this system on a random set of 100 \dataset{} questions shows 64.9 F1.} 
Table \ref{tab:type2} shows the distribution of the types of question. We also show distributions of other open-ended question datasets for comparison. 

In all datasets, concept questions are the most frequent; those questions ask about the definition or background knowledge.
Compared to other datasets, ours contain many more causal questions (e.g., \emph{why did Joyce Benes stop feeding the horses?}), reflecting that annotators frequently make causal inferences across events. In contrast we see fewer procedural and example questions. Our dataset also tend to contain few judgmental questions, i.e., question about opinions, which may be a reflection of news articles trying to stay objective.

\begin{table}[h]
\centering
\small
\begin{tabular}{c|rrr}
\toprule
  &INQ.  &Ours&TED-Q  \\
  \midrule
 verification&4.0 &7.9&15.5\\
 disjunctive&0.1 &1.0&1.3\\
 concept&31.3 &32.5&23.3\\
 extent&7.7 &5.7&4.9\\
 example& 13.7&6.9&15.0\\
 comparison& 0.6&0.5&0.8\\
 cause& 14.1&31.8&13.8\\
 consequence& 4.2&0.6&1.5\\
 procedural&14.3 &10.8&14.7\\
 judgmental& 9.9&2.4&9.2\\
  \bottomrule
\end{tabular}
\caption{Question types in each dataset, classified using the model from \citet{cao}. Our dataset has good coverage of the key question types in the other two datasets. }
\label{tab:type2}
\end{table}

\section{Additional model analysis}\label{app:analysis}

\subsection{Model accuracy by question type}\label{app:analysis:qtype}
Table \ref{tab:type} shows the accuracy stratified by different types of questions as classified using the model from \citet{cao}. The QA model performs well on extent and consequence questions, followed by concept, verification, and disjunct questions, while performing relatively worse on comparison or cause questions.
\begin{table}[h]
\centering
\small
\begin{tabular}{c|c}
\toprule
 Type &Accuracy   \\
  \midrule
concept &66.8\\
 verification&67.8 \\
 procedural&58.4 \\
 comparison&50.0 \\
cause &55.7\\
judgmental &62.6 \\
 extent&71.3 \\
 example &64.3 \\
disjunct &66.7 \\
consequence &69.0 \\
  \bottomrule
\end{tabular}
\caption{Model performance statified by question type, trained and tested on DCQA.}
\label{tab:type}
\end{table}

\section{Pretraining details}
\label{sec:pretrain}
\subsection{SQuAD pretraining}\label{app:pretrain:squad}
At a high level, certain phrases in the SQuAD questions may be referred to by pronouns or demonstratives in their corresponding article, and we can replace such phrases in the questions by these more context-dependent forms, and sample an element in the chain that precedes the answer as the highlighted span.
Specifically, we:
\textbf{(1)} combine 5 consecutive paragraphs from SQuAD articles to create longer text and extract coreference chains;\footnote{We use the AllenNLP tool~\cite{Gardner2017AllenNLP}.}
\textbf{(2)} for each question whose answer span is in the combined text, we look for exact matches between ngrams in the question and expressions in coreference chains;
\textbf{(3)} if there is a match, substitute the question phrase with a random reference in the matched chain, or the demonstrative ``this''; When choosing the random reference we also consider whether the it is in possessive form.
\textbf{(4)} designate a random chain element preceding the answer as the highlighted span.

For example, for the SQuAD question ``\emph{Of what group in the periodic table is oxygen a member?}", we found ``oxygen'' in the corresponding article referred to as ``it'', ``the element'', etc. Thus we change the question to \emph{``Of what group in the periodic table is \textbf{it} a member?}", and generate the following highlight using one of the sentences before the answer: \emph{``At standard temperature and pressure, two atoms of \underline{the element} bind to form dioxygen.''}

This method synthesizes questions with noun spans only, which is the most frequent span category in {\sc inquisitive}; we did not handle verbs or adjectives this way because of their variability.

We synthesized 45437 questions, including 42021 questions with an answer. We use special tokens to denote the start and end of the synthesized span.

\subsection{ELI5 pretraining}\label{app:pretrain:eli5}

We retrieve the sentence in the supporting documents with the highest BM25 score as the approximate answer sentence.  Following \citet{KILT}, we use the whole English Wikipedia as the supporting corpus instead of the original supporting documents. We combine sentences before and after the answer sentence with the answer itself to form the ``article'' as the input to the question answering model. The number of context sentences are randomly chosen so that the length of the synthesized article is comparable to {\sc inquisitive} articles and the answers are evenly distributed among different positions in the synthesized article.

To prevent low quality answers, we have an additional filtering step that keeps only the examples whose cosine similarity between the sentence embeddings (embedded using distilled SRoBERTa~\cite{sbert}) of the retrieved answer and the gold answer is larger than 0.55. This resulted in 55740 examples.

\label{sec:instructions}
\begin{figure*}[t]
  \centering
  \includegraphics[width=1	\textwidth]{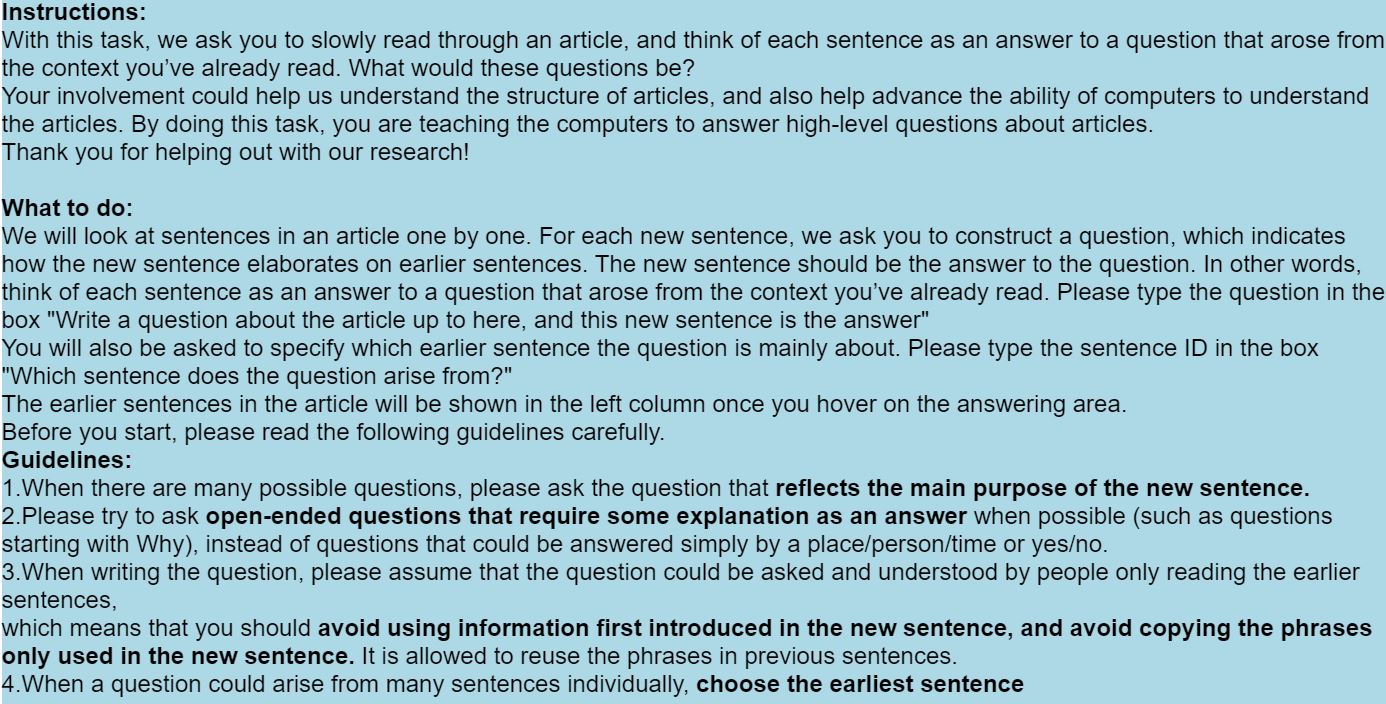}
  \caption{Key instructions of our crowdsourcing task for question collection.}
  \label{fig:instructions}
\end{figure*}

\subsection{Additional experimental details}
We run our experiments on NVIDIA Tesla V100 SXM2 GPUs. We use 2 GPUs when training the model and it takes about 10 hours for the QA model. The longformer-base-4096 we use has roughly 148M parameters. Our hyperparameters are tuned on the validation set; we mainly tune the learning rate and the number of epochs. Our reported results are from a single run.

\section{Instructions for question collection}
Instructions for \dataset{}'s crowdsourcing interface is shown in Figure~\ref{fig:instructions}.

\section{Copyright information related to \dataset{}}\label{app:copyright}
\dataset{}'s annotated data is a subset of the articles used in the {\sc Inquisitive} dataset~\cite{inq}. This data is sourced from three existing datasets: Newsela~\cite{newsela}, WSJ articles from the Penn Treebank~\cite{ptb}, and Associated Press~\cite{ap}. The Newsela dataset can be requested free of charge for researchers at \url{https://newsela.com/data}, and the authors have obtained permission to perform research on this data. The Penn Treebank is one of the most widely used resource in NLP, and is available via the LDC at \url{https://catalog.ldc.upenn.edu/LDC99T42}. The Associated Press data is also available from the LDC at \url{https://catalog.ldc.upenn.edu/LDC93T3A}.

\end{document}